\documentclass[10pt,twocolumn,letterpaper]{article}

\usepackage[pagenumbers]{cvpr} 

\definecolor{cvprblue}{rgb}{0.21,0.49,0.74}
\usepackage[pagebackref,breaklinks,colorlinks,allcolors=cvprblue]{hyperref}
\usepackage{adjustbox}
\usepackage{booktabs}
\usepackage{siunitx}
\usepackage{colortbl}   
\usepackage[table]{xcolor} 
\usepackage{adjustbox} 
\usepackage{subcaption}
\usepackage{multirow}
\definecolor{gray15}{gray}{0.94} 
\def\paperID{****} 
\def\confName{CVPR}
\def\confYear{2026}

\title{DenoiseGS: Gaussian Reconstruction Model for Burst Denoising}

\author{Yongsen Cheng$^{*, 1}$, Yuanhao Cai\thanks{Equal Contribution. $\dagger$ Corresponding Author: Yulun Zhang.}$^{*, 2}$, Yulun Zhang$^{\dagger,1}$\\
	$^1$ Shanghai Jiao Tong University,\quad 
    $^2$ Johns Hopkins University
}

\begin{document}
\maketitle

\begin{abstract}

\noindent Burst denoising methods are crucial for enhancing images captured on handheld devices, but they often struggle with large motion or suffer from prohibitive computational costs. In this paper, we propose DenoiseGS, the first framework to leverage the efficiency of 3D Gaussian Splatting for burst denoising. Our approach addresses two key challenges when applying feedforward Gaussian reconsturction model to noisy inputs: the degradation of Gaussian point clouds and the loss of fine details. To this end, we propose a Gaussian self-consistency (GSC) loss, which regularizes the geometry predicted from noisy inputs with high-quality Gaussian point clouds. These point clouds are generated from clean inputs by the same model that we are training, thereby alleviating potential bias or domain gaps. Additionally, we introduce a log-weighted frequency (LWF) loss to strengthen supervision within the spectral domain, effectively preserving fine-grained details. The LWF loss adaptively weights frequency discrepancies in a logarithmic manner, emphasizing challenging high-frequency details. Extensive experiments demonstrate that DenoiseGS significantly exceeds the state-of-the-art NeRF-based methods on both burst denoising and novel view synthesis under noisy conditions, while achieving \textbf{250$\times$} faster inference speed. Code and models are released at  \url{https://github.com/yscheng04/DenoiseGS}.
\end{abstract}

\begin{figure} [t]
    \centering \vspace{-2pt}
    \includegraphics[width=0.8\columnwidth]{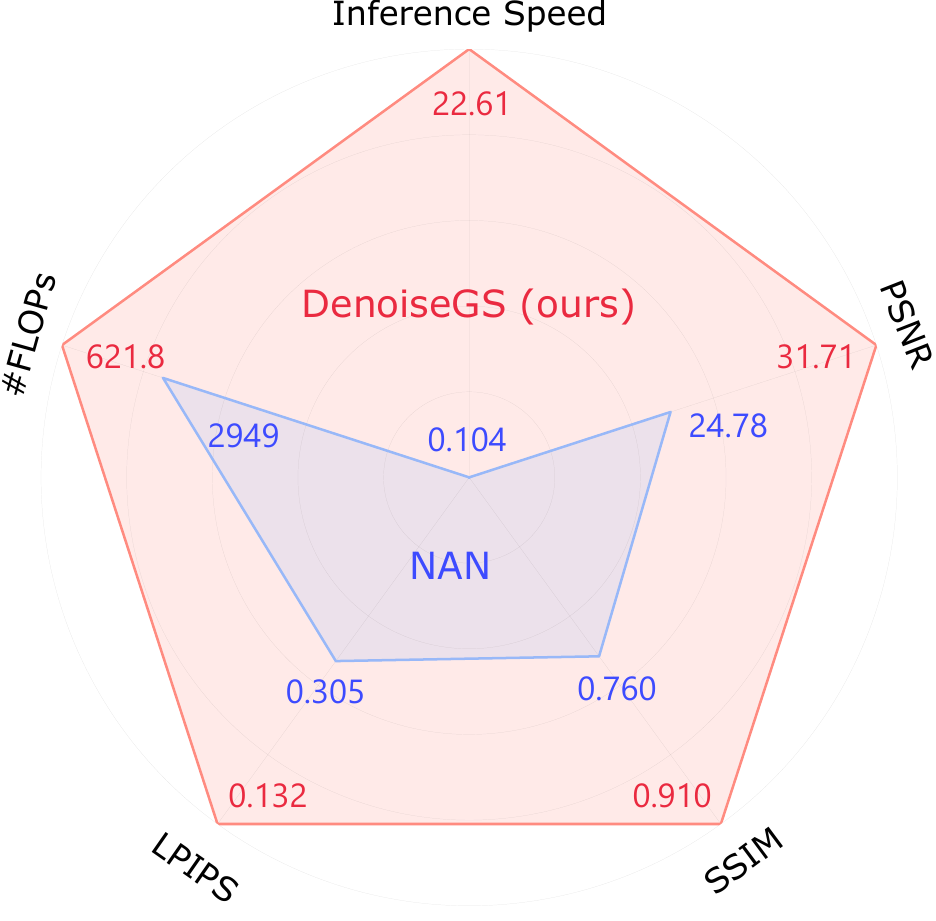}
    \vspace{-3mm}
    \caption{DenoiseGS $vs.$ NAN~\citep{pearl2022nan}. Our DenoiseGS achieves better PSNR, SSIM, LPIPS values with less FLOPs (in GFLOPs) and faster inference speed (in FPS).}
    \vspace{-7mm}
    \label{fig:comparison_nan_denoiseGS}
\end{figure}

\setlength{\abovedisplayskip}{2pt}
\setlength{\belowdisplayskip}{2pt}

\vspace{-3mm}
\section{Introduction}
With the growing popularity of smartphones, casual photography has become ubiquitous. However, images captured by handheld devices often suffer from noise due to limited sensor size and challenging imaging conditions. Image denoising, therefore, plays a crucial role in restoring clean results. Among existing approaches, burst denoising, which leverages multiple short-exposure frames, has proven particularly effective. It can exploit inter-frame redundancy to recover details that are lost in single-frame denoising. Nevertheless, burst denoising remains challenging because of complex motion and misalignment across frames.

Traditional 2D burst denoising methods~\citep{bhat2021deep, xia2020basis, dudhane2022burst} typically rely on accurate frame alignment or optical flow estimation, which can easily fail under large motion. On the other hand, modern 3D modeling approaches, exemplified by Neural Radiance Fields (NeRF)~\citep{mildenhall2021nerf} offers a more geometrically consistent solution by modeling the 3D structure of a scene. Compared with traditional 2D methods, NeRF-based approaches inherently handle complex parallax and large camera motions, allowing multi-view information to be fused in a physically consistent manner. Leveraging this advantage, Noise-Aware NeRFs (NAN)~\citep{pearl2022nan} introduces NeRF into burst denoising. By extending the IBRNet~\citep{wang2021ibrnet} with noise-aware components, NAN~\citep{pearl2022nan} achieves impressive denoising performance under challenging motions and heavy noise. However, NeRF relies on computationally expensive volume rendering, leading to slow inference speed. For instance, NAN takes 9.6s to denoise a single image at the resolution of 256$\times$256. This drawback makes it difficult to apply NeRF-based models to practical restoration tasks.

The recent advent of 3D Gaussian Splatting (3DGS)~\citep{kerbl20233d} offers a high-speed solution for efficiently tackling 3D reconstruction and NVS challenges. While vanilla 3DGS requires per-scene optimization, feed-forward Gaussian models~\citep{zhang2024gs, chen2024mvsplat, charatan2024pixelsplat} have enabled efficient inference without test-time fitting. For example, GS-LRM~\citep{zhang2024gs}, one of these SOTA feed-forward Gaussian reconstruction models, can efficiently produce novel views in less than 0.1s with as few as 2 input images. The efficiency of these feed-forward methods makes them a promising solution to burst denoising under large motions and high noise levels. However, directly applying these feedforward 3DGS models to burst denoising may encounter two key issues. \textbf{First}, they struggle to generate high-quality Gaussian point clouds from noisy inputs. \textbf{Second}, fine details are often lost in the rendered results, especially under high noise levels.

To address the above problems, we propose DenoiseGS, an efficient framework for burst denoising built on GS-LRM~\citep{zhang2024gs}. Our framework adopts two key components designed to enhance both the spatial structure of reconstructed Gaussian point clouds and the detail preservation of rendered images. \textbf{First}, the Gaussian self-consistency (GSC) loss regularizes the geometry with model's own predictions. During training, we additionally feed clean image bursts into the model to generate Gaussian point clouds with higher quality. The generated Gaussian point clouds can then serve as high-quality 3D guidance. \textbf{Second}, the log-weighted frequency (LWF) loss complements spatial supervision with frequency-domain constraints. The LWF loss adaptively weights frequency discrepancies in a logarithmic manner. This design emphasizes challenging high-frequency signals, encouraging the model to better preserve fine details. Extensive experiments demonstrate that our approach enhances GS-LRM~\citep{zhang2024gs} without introducing extra overhead and significantly surpasses SOTA burst denoising methods, while achieving faster inference. We also extend our model to novel view synthesis task under noisy conditions, reaching SOTA performance. Overall, the main contributions of our work are outlined below:

\begin{itemize}
    \vspace{2mm}
    \item We propose a novel framework, DenoiseGS, for burst denoising. To the best of our knowledge, this paper is the first to introduce 3D Gaussian splatting into the task of burst denoising.
    \vspace{2mm}
    \item We design a Gaussian self-consistency (GSC) loss that exploits the model’s inherent ability to produce high-quality Gaussian point clouds from clean input to enhance the quality of point clouds reconstructed from noisy ones.
    \vspace{-2mm}
    \item We propose a log-weighted frequency (LWF) loss to better preserve fine details in reconstructed images.
    \vspace{2mm}
    \item Extensive experiments demonstrate that our method surpasses SOTA methods in both burst denoising and novel view synthesis, while achieving faster inference.
\end{itemize}

\section{Related Work}
\textbf{Neural Radiance Field.} NeRF~\citep{mildenhall2021nerf} models scenes via a continuous implicit function to predict emitted radiance and volume density from a point's coordinates and viewing angle. It shows remarkable success in the novel view synthesis (NVS) task, motivating numerous subsequent studies that aim to enhance reconstruction fidelity~\citep{barron2021mip, barron2022mip, verbin2022ref, hu2023tri}, accelerate inference~\citep{muller2022instant, chen2023mobilenerf, reiser2023merf}, or broaden its range of applications~\citep{pearl2022nan, cai2024structure, cui2024aleth}. As an example, \citet{pearl2022nan} introduces Noise-Aware NeRF, which augments IBRNet~\citep{wang2021ibrnet} with additional noise-aware components to help reconstruct scene structures from noisy bursts of images. While these approaches yield promising results, NeRF-based methods still face significant limitations in both training and inference efficiency, primarily due to the heavy computational burden introduced by the volume rendering.

\textbf{Gaussian Splatting.} 3DGS~\citep{kerbl20233d} constructs explicit scene representations comprised of a vast number of Gaussian primitives. By adopting differentiable rasterization, it attains superior rendering speeds relative to NeRF-based methods, which are bottlenecked by computationally expensive volumetric rendering. Consequently, 3DGS has quickly gained popularity and found applications in various domains, including dynamic scene rendering~\citep{yang2023real, wu20244d, luiten2024dynamic}, SLAM~\citep{keetha2024splatam, yugay2023gaussian, matsuki2024gaussian, yan2024gs}, inverse rendering~\citep{liang2024gs, xie2024physgaussian, jiang2024gaussianshader}, digital humans~\citep{liu2024humangaussian, kocabas2024hugs, hu2024gauhuman}, 3D content generation~\citep{tang2023dreamgaussian, yi2024gaussiandreamer, liang2024luciddreamer}, and medical imaging~\citep{cai2024radiative, zha2024r}. Traditional 3DGS methods typically require test-time optimization for each individual scene. In contrast, recent feed-forward models~\citep{charatan2024pixelsplat, chen2024mvsplat, xu2025depthsplat, zhang2024gs} can reconstruct a scene in a single forward pass using as few as two images, thereby achieving significantly faster inference speeds. We leverage one of these models, GS-LRM~\citep{zhang2024gs}, as our base model for burst denoising.

\textbf{Burst Denoising.} Early burst denoising techniques recover clean target images by predicting per-input-image per-pixel denoising kernels~\citep{mildenhall2018burst} or employing a Lucas-Kanade tracker to find a homography for each frame before denoising~\citep{godard2018deep}. These methods yield promising results, but they can only deal with motions up to 2 pixels and show limited denoising capability. To address these issues, \citet{xia2020basis} uses larger denoising kernels to aggregate more spatial information. \citet{bhat2021deep} aligns each input image with the target images with optical flow and then denoises the images in a deep feature space. \citet{dudhane2022burst} proposes an edge-boosting feature alignment module and a pseudo-burst feature aggregation module. However, these 2D methods still lack 3D perception capabilities. Inspired by the success of 3D reconstruction techniques, researchers have started using 3D reconstruction models for burst denoising. \citet{pearl2022nan} extends the IBRNet~\citep{wang2021ibrnet} structure to handle noise. \citet{tanay2023efficient} extends the multiplane image framework and introduces multiplane feature representation. Yet, these methods are either limited by high computational cost or by the capabilities of the 3D representation.

\textbf{Novel View Synthesis in Degraded Scenes.} While vanilla 3DGS~\citep{kerbl20233d} requires high-quality images as input, degraded images are common in many real-world scenarios. Several Nerf- and 3DGS-based methods~\citep{pearl2022nan, zhou2023nerflix, zhou2023nerflix++, feng2024srgs, linhqgs} synthesize novel view from degraded input images. \citet{pearl2022nan} proposed a noise-aware NeRF framework originally designed for burst denoising tasks, but is applicable to novel view synthesis from noisy input images. Rather than focus on 3D learning, NeRFLiX~\citep{zhou2023nerflix} and NeRFLiX++~\citep{zhou2023nerflix++} restore the rendered outputs with a postprocessing module Inter-Viewpoint Mixer. Recently, SRGS~\citep{feng2024srgs} was proposed to generate high-resolution renderings from low-resolution inputs using 3DGS. To this end, the method incorporates a pretrained 2D super-resolution model into its pipeline. HQGS~\citep{linhqgs} takes a different approach and improves the quality of the reconstructed point cloud by enhancing degraded inputs with high-frequency, edge-aware maps.

\section{Problem Setup}

Given a burst of noisy images, the task of burst denoising includes aggregating information from input images and then denoising one of them as the result. In this work, the target image to denoise is randomly chosen from the input images. We also choose to formulate the input as an unordered set rather than a sequential stream, primarily due to the potential motion and misalignment between consecutive frames.

Given that our model is built upon GS-LRM~\citep{zhang2024gs}, a framework originally designed for novel view synthesis, it is natural to extend its capabilities to noisy conditions. Consequently, we further train and evaluate our method on the task of synthesizing novel views from noisy inputs.

\begin{figure} [t]
    \centering 
    \vspace{1mm}
    \includegraphics[width=1.0\columnwidth]{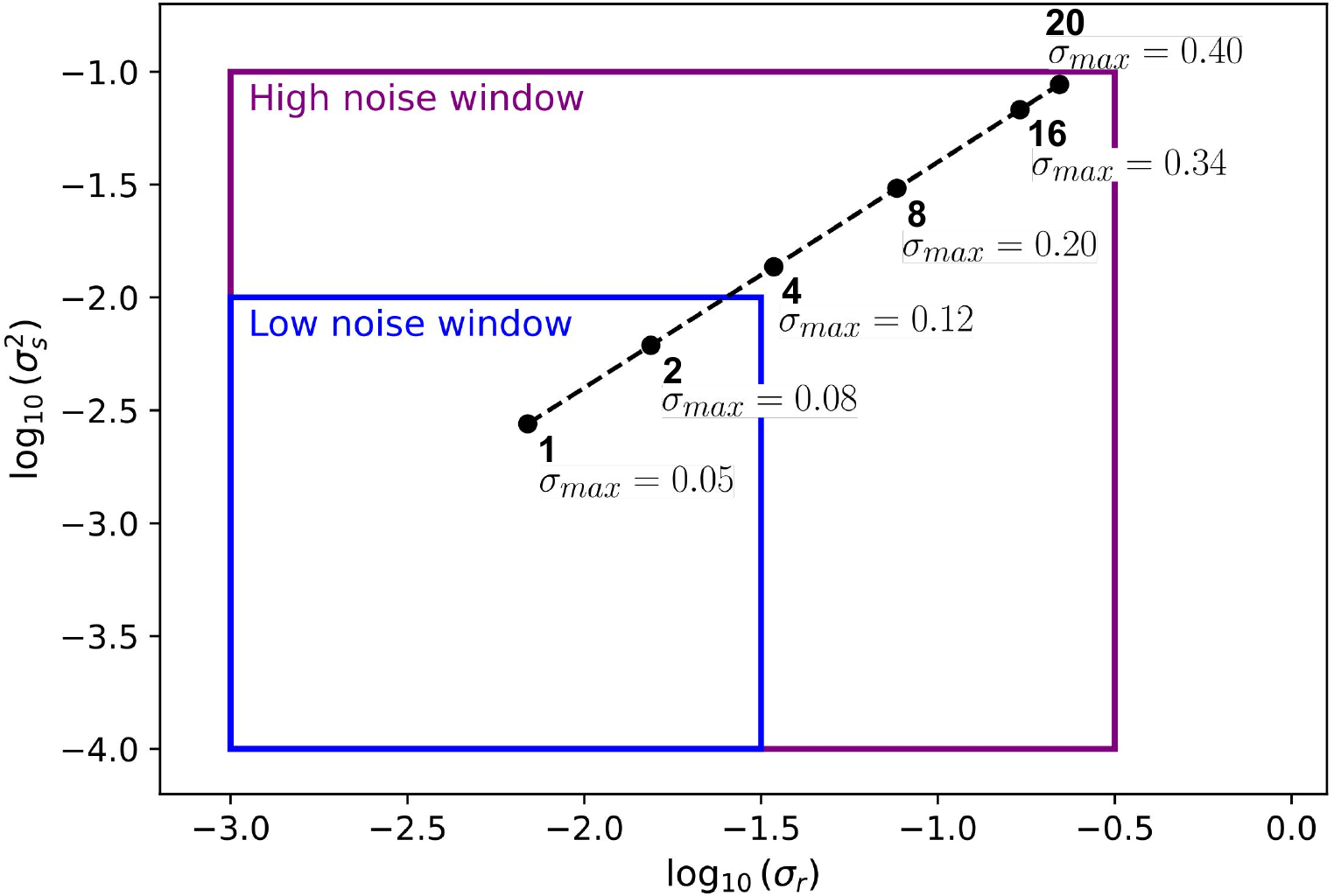}
    \vspace{-7mm}
    \caption{Noise parameters used during training and evaluation. Values (1-20) beside the point indicate gain levels defined in KPN~\citep{mildenhall2018burst}. $\sigma_{\max} = \sqrt{\sigma_r^2 + \sigma_s^2}$ represents the maximum noise.
    } \vspace{-5mm}
    \label{fig:gain_levels}
\end{figure}

Following the practice in KPN~\citep{mildenhall2018burst}, we adopt a noise model that formulates the relationship between the noisy version and its clean linear counterpart as follows:
\begin{equation}
\label{noise}
I_n(p) \sim \mathcal{N}(I_n^c(p), \sigma_r^2 + \sigma_s^2I_n^c(p)),
\end{equation}
where $I_n(p), I_n^c(p)$ refer to the intensity of the noisy and clean images at pixel $p$. $\sigma_r$ and $\sigma_s$ are noise parameters that depend on the sensor gain level (i.e., ISO). $\mathcal{N}$ represents the Gaussian distribution. Following KPN~\citep{mildenhall2018burst}, we evaluate our model over different gain levels of a sample camera. The relationship between gain levels and noise parameters $\sigma_r, \sigma_s$ is illustrated in Fig.~\ref{fig:gain_levels}. We train our model in the high noise window (purple rectangle in Fig.~\ref{fig:gain_levels}) and evaluate our model on gain levels (black points in Fig.~\ref{fig:gain_levels}) up to 20.

\section{Methods}

\begin{figure*} [t]
    \centering 
    \includegraphics[width=1.0\textwidth]{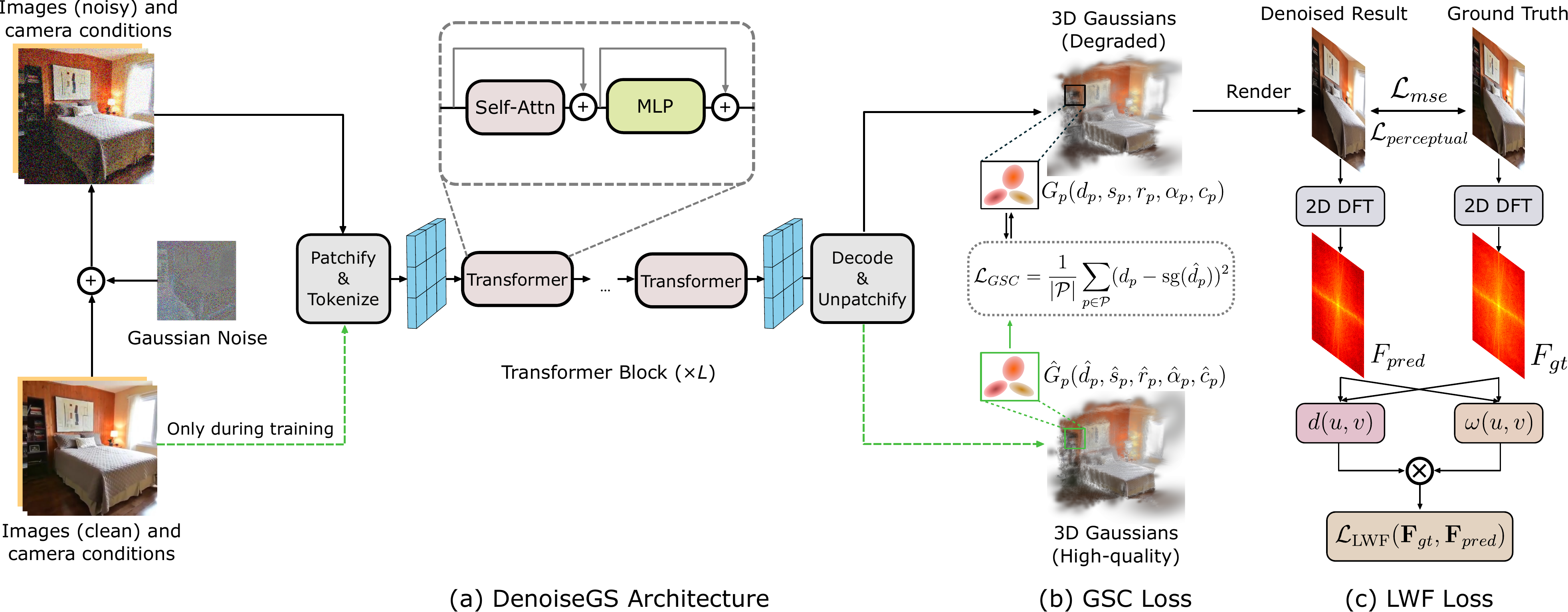}\vspace{-0.2cm}
    \caption{Pipeline. (a) Multi-view noisy inputs and camera conditions are processed by Transformer blocks to predict per-pixel Gaussian. During training, clean inputs are also fed into the model to generate high-quality Gaussian point clouds. (b) The high-quality point cloud is then used as guidance for the point cloud reconstructed from noisy inputs with our proposed Gaussian self-consistency loss. (c) The denoised result is further supervised in frequency domain with our proposed log-weighted frequency loss.} 
    \vspace{-5mm}
    \label{fig:pipeline}
\end{figure*}

Figure~\ref{fig:pipeline} illustrates the overall framework of our method. We begin by retraining GS-LRM~\citep{zhang2024gs} on our customized 3D noisy dataset RE10K-N as a simple baseline. The RE10K-N is adapted from the large-scale 3D dataset RE10K\citep{zhou2018stereo}. To enhance the quality of the reconstructed Gaussian point clouds, we then propose the Gaussian self-consistency (GSC) loss. Specifically, clean image bursts are simultaneously fed into the model during training to generate high-quality Gaussian point clouds. These point clouds then serve as 3D guidance for the point clouds reconstructed from the corresponding noisy bursts. Subsequently, to better capture relative depth and geometry, we replace the original Pl\"{u}cker rays~\citep{plucker1865xvii} used in GS-LRM with the more effective reference-point Pl\"{u}cker coordinate (RPPC)~\citep{cai2025baking}. Furthermore, we propose the log-weighted frequency (LWF) loss as supplementary frequency-level supervision, encouraging the model to preserve more high-frequency details.

\subsection{A Simple Baseline}
\label{sec:baseline}

We build our model upon GS-LRM~\citep{zhang2024gs}, a large reconstruction model for Gaussian splatting with strong spatial perception capability. We adopt GS-LRM as our baseline primarily because it is a feedforward 3DGS model that requires no per-scene optimization, enabling extremely fast inference. Moreover, it can reconstruct high-quality Gaussian point clouds from as few as two input images, demonstrating its strong reconstruction ability.

\noindent\textbf{Simple Baseline.} We retrain GS-LRM on the burst denoising task as a simple baseline. As illustrated in Fig.~\ref{fig:pipeline} (a), the multi-view inputs are first concatenated with camera conditions, and then are embedded and processed with standard Transformer blocks. Following the Transformer blocks, the network outputs per-pixel Gaussian parameters that define the 3D Gaussian point cloud. Specifically, each Gaussian $G_p$ corresponding to pixel $p$ is parameterized by a depth $d_p \in \mathbb{R}$, a scale $s_p \in \mathbb{R}^3$, a rotation quaternion $r_p \in \mathbb{R}^4$, an opacity $\alpha_p \in \mathbb{R}$, and an RGB color $c_p \in \mathbb{R}^3$.
Given the camera origin $\mathbf{o}$ and the ray direction $\mathbf{d}_p$ of the pixel-aligned ray associated with pixel $p$, the 3D position of the Gaussian center is then computed as:
\begin{equation}
    \boldsymbol{\mu}_p = \mathbf{o} + d_p \mathbf{d}_p.
\end{equation}
As the final step, the reconstructed Gaussians are rendered through a standard differentiable 3DGS rasterization pipeline to produce images of target views.

\noindent\textbf{RE10K-N.} Since GS-LRM tends to overfit when trained on the LLFF-N~\citep{pearl2022nan} dataset proposed in NAN, we construct a new dataset RE10K-N for training and evaluation based on the large-scale RealEstate10K (RE10K) dataset~\citep{zhou2018stereo}, which was used by GS-LRM~\citep{zhang2024gs}. Specifically, we take clean scene images from RE10K and synthesize noisy counterparts following the noise model defined in Eq.~\eqref{noise}.

Before adding noise, as the model operates in the linear RGB space, we ``linearize'' the images from RE10K by reversing the gamma correction and applying inverse white balancing. After adding noise, instead of directly using the linearized images as inputs as in previous burst denoising methods~\citep{mildenhall2018burst, xia2020basis, bhat2021deep, pearl2022nan}, we ``delinearize'' them back to sRGB and feed these processed images to the model. Since delinearization does not require any additional information, it does not compromise fairness in comparison.

\noindent\textbf{Limitations of GS-LRM.} The retrained GS-LRM achieves competitive performance. However, we observe two main limitations. \textbf{First}, since the final result is rendered from the reconstructed Gaussian point clouds, the quality of the point clouds critically affect the denoising results. However, as is shown in Fig.~\ref{fig:depth-map} (c) and Fig.~\ref{fig:depth-map} (d), the depth map predicted under noisy conditions differs greatly from the clean one, indicating point clouds tend to degrade under noisy conditions. \textbf{Second}, the denoised images exhibit noticeable loss of fine details, particularly at high gain levels.

In the following sections, we will present our methods designed to ameliorate these issues and enhance GS-LRM’s performance in point cloud quality and detail preservation.

\subsection{Gaussian Self-Consistency Loss}

\begin{figure} [t]
    \centering 
    \includegraphics[width=1.0\columnwidth]{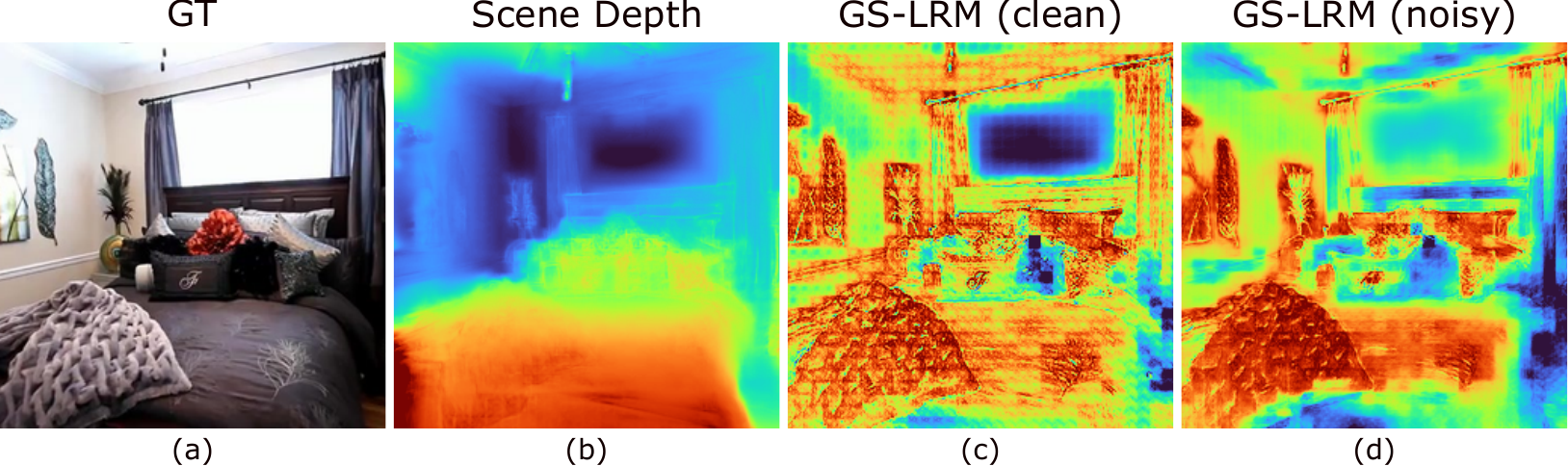}
    \vspace{-7mm}
    \caption{From left to right: (a) clean image; (b) scene depth (c) depth map predicted by GS-LRM on clean inputs; (d) depth map predicted by GS-LRM on noisy inputs.} 
    \vspace{-5mm}
    \label{fig:depth-map}
\end{figure}

\textbf{Motivation.} To enhance the quality of the generated Gaussian point clouds, a potential solution is to incorporate additional depth priors, as in DepthSplat~\citep{xu2025depthsplat}. However, this approach has two inherent limitations. First, it relies on an external pretrained depth estimation network to predict pixel-wise depth maps, inevitably introducing its bias into the training process. Second, as our model is trained solely for burst denoising instead of novel view synthesis, the depth maps rendered from Gaussian point clouds do not align with the true scene depth, as is illustrated in Fig.~\ref{fig:depth-map}. The depth map predicted by GS-LRM in Fig.~\ref{fig:depth-map} (c) exhibits a pattern that differs notably from the scene depth in Fig.~\ref{fig:depth-map} (b). These patterns are learned to enhance visual quality from specific viewpoints rather than to recover accurate geometry. Therefore, forcing our model's depth map to align with an externally predicted depth map per-pixel may impair the reconstruction capacity of the model.

Instead, we can take advantage of the model’s inherent ability to generalize across different noise levels. Since the model is trained with bursts of varying noise levels, it naturally learns to handle both noisy and clean inputs, and can predict higher-quality point clouds from clean inputs. Therefore, Gaussian point clouds reconstructed from clean bursts can serve as reliable supervision for those generated from noisy ones. This enables the model to refine the Gaussian point clouds without requiring external depth priors.

\noindent \textbf{GSC Loss.} Inspired by the above observations, we propose the Gaussian self-consistency loss. It regularizes the Gaussian point clouds predicted from noisy images with the model’s own predictions from clean inputs. Specifically, during training, we additionally feed clean images into the model to generate higher-quality Gaussian point clouds. These point clouds then serve as 3D guidance for those reconstructed from noisy images. Let the Gaussian point clouds predicted from noisy images be denoted as:
\begin{equation}
\mathcal{G}=\{G_p(d_p, s_p, r_p, \alpha_p, c_p) | p\in \mathcal{P} \},
\end{equation}
where $\mathcal P$ denotes all the pixels in the input images and $G_p$ denotes the Gaussian associated with the pixel $p$. Its depth, scale, rotation, opacity and color are denoted as $d_p \in \mathbb{R}$, $s_p \in \mathbb{R}^3$, $r_p \in \mathbb{R}^4$, $\alpha_p \in \mathbb R$, $c_p \in \mathbb R^3$. The Gaussian point clouds predicted from clean images is denoted as:
\begin{equation}
\label{eq:clean_gaussians}
\mathcal{G}_{clean}=\{\hat G_p(\hat d_p, \hat s_p, \hat r_p, \hat \alpha_p, \hat c_p) | p\in \mathcal{P} \}.
\end{equation}

As the model outputs per-pixel Gaussians, a natural one-to-one correspondence can be established between the Gaussians predicted from clean and noisy inputs. Based on this correspondence, we define the GSC Loss as the L2 loss between the depth attributes of these two Gaussian sets:
\begin{equation}
\label{eq:GSC}
\mathcal{L}_{GSC}(\mathcal G, \mathcal G_{clean}) = \frac 1 {|\mathcal{P}|} \sum_{p\in \mathcal P} (d_p - \text{sg}(\hat d_p))^2,
\end{equation}
where $|\mathcal P|$ denotes the number of pixels in the input images, which is also equal to the total number of the Gaussians. The $\text{sg}(\cdot)$ denotes the stop-gradient operation, which prevents gradients from flowing through the clean branch. The stop-gradient operation ensures that the model learns to make the Gaussian point clouds generated under noisy conditions approach those from clean conditions, rather than the other way around. It is worth noting that we only use the depth attribute in this loss, without involving other Gaussian properties. Our experiments show that adding multiple attributes to the loss may impair the training stability. Among these Gaussian attributes, depth is the most crucial factor for preserving the structure of the point cloud.

\noindent \textbf{Warm-up Phase.} To ensure the effectiveness of our GSC loss, it is crucial that the model is capable of reconstructing higher-quality Gaussians from clean images than from noisy ones — a condition not guaranteed at the early stage of training. Therefore, to ensure the guidance is meaningful, we introduce a warm-up phase, during which the GSC loss is not applied until the model reaches a sufficient reconstruction quality on clean inputs.

\noindent\textbf{Reference-Point Plücker Coordinate.}
To provide the model with a more spatially informative camera condition, 
we adopt the RPPC proposed in \citep{cai2025baking} to replace the conventional Plücker ray representation. Existing approaches typically formulate pixel-aligned Plücker coordinates as \(\mathbf{r} = (\mathbf{o} \times \mathbf{d}, \mathbf{d})\), with \(\mathbf{o}\) and \(\mathbf{d}\) representing the ray's origin and direction, respectively. Nevertheless, the moment vector \(\mathbf{o} \times \mathbf{d}\) primarily encodes angular information, which hinders the effective extraction of scene geometry and relative depth.

To address this, RPPC adopts a reference point located at the ray's closest approach to the origin, instead of using the moment vector. Formally, this is expressed as:
\begin{equation}
    \mathbf{r} = (\mathbf{o} - (\mathbf{o} \cdot \mathbf{d})\mathbf{d}, \mathbf{d}).
\end{equation}
Compared to the standard Plücker coordinates, RPPC provides a more geometry-aware conditioning signal for the model, allowing it to better perceive the camera’s spatial conditions and more effectively interpret the 3D geometric relationships among multi-view inputs.

\vspace{-1mm}
\subsection{Log-Weighted Frequency Loss}
\vspace{-1mm}
To encourage the model to reconstruct high-frequency details more accurately, 
we employ a log-weighted frequency (LWF) loss in the frequency domain. 
Given the ground-truth and predicted images $I_{gt}$ and $I_{pred}$, 
their discrete Fourier transforms are defined as:
\begin{equation}
\begin{split}
\mathbf{F}_{gt}(u,v) &= 
\sum_{h=0}^{H-1} \sum_{w=0}^{W-1} 
I_{gt}(h,w)
e^{-j2\pi(\frac{uh}{H}+\frac{vw}{W})}, \\
\mathbf{F}_{pred}(u,v) &= 
\sum_{h=0}^{H-1} \sum_{w=0}^{W-1} 
I_{pred}(h,w)
e^{-j2\pi(\frac{uh}{H}+\frac{vw}{W})},
\end{split}
\label{eq:fft_def}
\end{equation}
where $\mathbf{F}_{gt}, \mathbf{F}_{pred} \in \mathbb{R}^{H\times W \times C}$ are the frequency spectrum of all channels corresponding to $I_{gt}$ and $I_{pred}$. The magnitude difference at frequency location $(u,v)$ is given by
\begin{equation}
d(u,v) = \| \mathbf{F}_{gt}(u,v) - \mathbf{F}_{pred}(u,v) \|.
\end{equation}
We then introduce a logarithmic weighting term:
\begin{equation}
\omega(u,v) = \log(\sqrt{d(u,v)} + 1),
\end{equation}
which adaptively emphasizes harder frequency components with larger reconstruction errors. 
The final Log-weighted Frequency Loss is formulated as:
\begin{equation}
\mathcal L_{\text{LWF}}(\mathbf{F}_{gt}, \mathbf{F}_{pred}) 
= \frac{1}{HW} \sum_{u=0}^{H-1} \sum_{v=0}^{W-1} 
\omega(u,v) \, d(u,v).
\label{eq:log_freq_loss}
\end{equation}

This formulation balances the frequency-domain supervision by 
focusing more on challenging frequency regions while maintaining training stability.

\setlength{\tabcolsep}{1pt}
\begin{table*}[t]
\centering
\begin{adjustbox}{width=\textwidth}
\begin{tabular}{@{}l c ccccccccccccccccccc@{}}
\toprule
\multirow{2}{*}{\vspace{-2mm} Method} & Inference Speed & \multicolumn{3}{c}{Gain 1} & \multicolumn{3}{c}{Gain 2} & \multicolumn{3}{c}{Gain 4} & \multicolumn{3}{c}{Gain 8} & \multicolumn{3}{c}{Gain 16} & \multicolumn{3}{c}{Gain 20} \\
\cmidrule(lr){2-2}
\cmidrule(lr){3-5}\cmidrule(lr){6-8}\cmidrule(lr){9-11}\cmidrule(lr){12-14}\cmidrule(lr){15-17}\cmidrule(lr){18-20}
& FPS $\uparrow$ & PSNR$\uparrow$ & SSIM$\uparrow$ & LPIPS$\downarrow$ & PSNR$\uparrow$ & SSIM$\uparrow$ & LPIPS$\downarrow$ & PSNR$\uparrow$ & SSIM$\uparrow$ & LPIPS$\downarrow$ & PSNR$\uparrow$ & SSIM$\uparrow$ & LPIPS$\downarrow$ & PSNR$\uparrow$ & SSIM$\uparrow$ & LPIPS$\downarrow$ & PSNR$\uparrow$ & SSIM$\uparrow$ & LPIPS$\downarrow$ \\
\midrule
\multicolumn{20}{c}{\emph{Burst Denoising}} \\
\midrule
\multicolumn{20}{c}{\textit{2D Methods}} \\
KPN~\citep{mildenhall2018burst} & \textbf{140.85} & 36.82 & 0.965 & 0.078 & 34.61 & 0.948 & 0.112 & 32.27 & 0.920 & 0.159 & 29.68 & 0.875 & 0.226 & 26.81 & 0.803 & 0.318 & 25.84 & 0.772 & 0.353 \\
BPN~\citep{xia2020basis} & 23.75 & 32.71 & 0.952 & 0.104 & 31.92 & 0.937 & 0.136 & 30.52 & 0.909 & 0.185 & 28.48 & 0.861 & 0.257 & 25.89 & 0.782 & 0.354 & 24.98 & 0.749 & 0.391 \\
Deeprep~\citep{bhat2021deep} & 20.49 & 37.18 & 0.968 & 0.059 & 35.92 & 0.958 & 0.070 & 33.80 & 0.937 & 0.097 & 31.11 & 0.898 & 0.145 & 28.18 & 0.839 & 0.216 & 27.23 & 0.815 & 0.243 \\
\multicolumn{20}{c}{\textit{3D Methods}} \\
NAN~\citep{pearl2022nan} & 0.10 & 24.27 & 0.835 & 0.222 & 24.43 & 0.828 & 0.232 & 24.67 & 0.810 & 0.255 & 24.78 & 0.760 & 0.305 & 23.68 & 0.637 & 0.399 & 22.76 & 0.573 & 0.440 \\
GS-LRM~\citep{zhang2024gs} & 21.86 & 37.01 & 0.970 & 0.049 & 35.42 & 0.958 & 0.068 & 33.39 & 0.938 & 0.097 & 30.87 & 0.904 & 0.140 & 27.93 & 0.848 & 0.203 & 26.80 & 0.822 & 0.231 \\
DenoiseGS & 22.61 & \textbf{37.98} & \textbf{0.972} & \textbf{0.046} & \textbf{36.34} & \textbf{0.961} & \textbf{0.065} & \textbf{34.22} & \textbf{0.942} & \textbf{0.092} & \textbf{31.71} & \textbf{0.910} & \textbf{0.132} & \textbf{28.74} & \textbf{0.856} & \textbf{0.191} & \textbf{27.66} & \textbf{0.832} & \textbf{0.216} \\
\midrule
\multicolumn{20}{c}{\emph{Novel View Synthesis Under Noisy Conditions}} \\
\midrule
NAN~\citep{pearl2022nan} & 0.10 & 23.56 & 0.773 & 0.260 & 23.74 & 0.769 & 0.267 & 24.04 & 0.758 & 0.284 & 24.37 & 0.723 & 0.320 & 23.79 & 0.624 & 0.394 & 23.08 & 0.571 & 0.429 \\
GS-LRM~\citep{zhang2024gs} & 21.86 & 25.34 & 0.839 & 0.146 & 25.24 & 0.834 & 0.154 & 25.02 & 0.825 & 0.170 & 24.55 & 0.804 & 0.198 & 23.59 & 0.763 & 0.249 & 23.10 & 0.742 & 0.273 \\
DenoiseGS & \textbf{22.61} & \textbf{25.97} & \textbf{0.854} & \textbf{0.134} & \textbf{25.86} & \textbf{0.849} & \textbf{0.142} & \textbf{25.62} & \textbf{0.839} & \textbf{0.158} & \textbf{25.12} & \textbf{0.818} & \textbf{0.185} & \textbf{24.14} & \textbf{0.779} & \textbf{0.234} & \textbf{23.64} & \textbf{0.760} & \textbf{0.254} \\
\bottomrule
\end{tabular}
\end{adjustbox}
\vspace{-3mm}
\caption{Quantitative results on RE10K-N dataset. Following NAN~\citep{pearl2022nan}, we evaluate our model on two distinct tasks. For burst denoising, the target image is part of the input sequence, whereas for novel view synthesis under noisy conditions, the target is intentionally excluded. GS-LRM and DenoiseGS are evaluated with a burst size of 2, while other methods use a burst size of 8. To ensure the fairness of comparison, the 2-frame bursts are always selected as subsets of the corresponding 8-frame bursts for each scene.  }
\vspace{-5mm}
\label{tab:quantitive_results}
\end{table*}

\subsection{Training Objective}
Following GS-LRM, we adopt the Mean Squared Error (MSE) loss and the Perceptual loss in our training objective. 
In addition, we incorporate the Gaussian Self-Consistency (GSC) loss defined in Eq.~\eqref{eq:GSC} 
and the Log-weighted Frequency (LWF) loss defined in Eq.~\eqref{eq:log_freq_loss}. 
Let $I_{pred}$ denote the rendered target and $I_{gt}$ denote the ground truth . 
The overall loss for our DenoiseGS is formulated as:
\begin{equation}
\begin{split}
\mathcal{L} = 
&~\lambda_{\text{mse}}\, \mathcal{L}_{\text{MSE}}(I_{pred}, I_{gt}) 
+ \lambda_{\text{lpips}}\, \mathcal{L}_{\text{LPIPS}}(I_{pred}, I_{gt}) \\
&+ \lambda_{\text{gsc}}\, \mathcal{L}_{\text{GSC}}(I_{pred}, I_{gt})
+ \lambda_{\text{freq}}\, \mathcal{L}_{\text{LWF}}(I_{pred}, I_{gt}).
\end{split}
\label{eq:Loss}
\end{equation}

\section{Experiments}

\vspace{-1mm}
\subsection{Experimental Settings}
\vspace{-1mm}
\textbf{Dataset.} We perform all of our main experiments on the RE10K-N dataset, with its generation process detailed in Sec.~\ref{sec:baseline}. Following prior work~\citep{charatan2024pixelsplat}, we adopt the standard training and testing split for fair comparison.

\noindent\textbf{Implementation Details.}
We implement DenoiseGS using the PyTorch framework~\citep{paszke2019pytorch} and train the model with the Adam optimizer~\citep{kingma2014adam} ($\beta_1$ = 0.9, $\beta_2$ = 0.95, and $\epsilon$ = 1$\times$10$^{-8}$). To accelerate both the training and inference phases, we additionally incorporate Flash-Attention~\citep{dao2023flashattention} from the xFormers library~\citep{lefaudeux2022xformers}, following practice in GS-LRM~\citep{zhang2024gs}. We fix the burst size to 2, as recent work~\citep{charatan2024pixelsplat} has shown this setting is generally sufficient for scene reconstruction and can balance reconstruction quality and efficiency. Further analysis on this trade-off of burst size is detailed in Sec.~\ref{sec:ablation}. The loss weights in Eq.~\eqref{eq:Loss} are set to $\lambda_{\text{mse}}$ = 1.0, $\lambda_{\text{lpips}}$ = 0.5, $\lambda_{\text{gsc}}$ = 0.06, and $\lambda_{\text{freq}}$ = 1.75. Our model is trained for 375k iterations on a single NVIDIA RTX A6000 GPU with images at 256$\times$256 resolution. The GSC loss is applied after the first 120k iterations to stabilize the initial training and ensure the GSC loss provides a meaningful and consistent guidance. Following NAN~\citep{pearl2022nan}, our model is trained on a high-noise range (highlighted with purple rectangle in Fig.~\ref{fig:gain_levels}) and evaluated across gain levels (black points in Fig.~\ref{fig:gain_levels}).

\vspace{-1mm}
\subsection{Comparison with Other Methods}
\vspace{-1mm}

\textbf{Quantitative Comparison on Burst Denoising.} We compare our method with a strong baseline (GS-LRM), state-of-the-art 2D methods (KPN~\citep{mildenhall2018burst}, BPN~\citep{xia2020basis}, Deeprep~\citep{bhat2021deep}), and the leading 3D method (NAN~\citep{pearl2022nan}), with quantitative and qualitative results reported in Tab.~\ref{tab:quantitive_results}. All competing baselines are retrained on our RE10K-N dataset over the high noise region (the purple rectangle in Fig.~\ref{fig:gain_levels}) to guarantee a fair comparison. It's worth noting that our method achieves superior results using a burst size of only 2, whereas competitors use 8, highlighting the efficiency of our approach. In terms of quantitative metrics (PSNR, SSIM, and LPIPS), our method consistently outperforms other baselines across all gain levels. Specifically, DenoiseGS is not only over 250$\times$ faster than NAN—which struggles to scale effectively on large datasets—but also delivers substantially better restoration quality. It also surpasses the leading 2D methods in denoising performance while maintaining fast inference speeds comparable to BPN and Deeprep. 

\begin{figure*} [t]
    \centering 
    \includegraphics[width=1.0\textwidth]{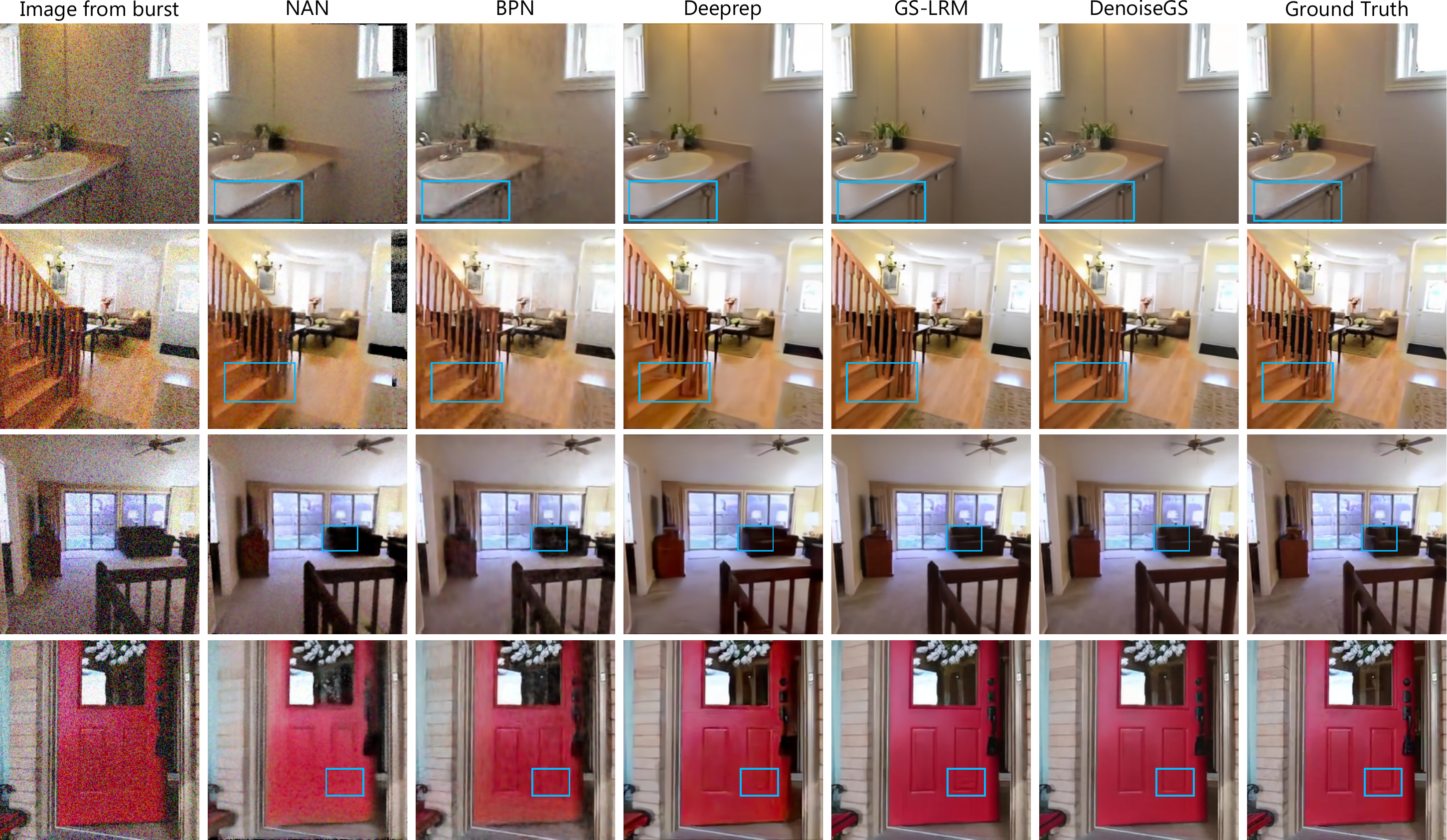}
    \vspace{-7mm}
    \caption{Qualitative comparison of burst denoising results on the RE10K-N dataset at gain 8. Key regions for comparison are highlighted with colored rectangles. While competing methods suffer from residual noise, over-smoothing, and visual artifacts, our approach robustly restores fine details and sharp structures from highly noisy inputs. Please zoom in for a better view.} 
    \vspace{-3mm}
    \label{fig:visual_burst_denoising}
\end{figure*}

\noindent \textbf{Qualitative Comparison on Burst Denoising.} These metric improvements aligned with the visual comparisons in Fig.~\ref{fig:visual_burst_denoising}, where competing methods exhibit several common failure modes such as residual noise (NAN, BPN), over-smoothing of textured regions (Deeprep), or visual artifacts (GS-LRM). In contrast, since our DenoiseGS is trained under additional 3D-level guidance and supplementary frequency-domain supervision provided by our proposed GSC and LWF losses, it can generate clean results with fine details from inputs of high noise levels.

\noindent\textbf{Quantitative Comparison on Novel View Synthesis.} The quantitative results of novel view synthesis on RE10K-N are reported in Tab.~\ref{tab:quantitive_results}. We compare our method with our baseline GS-LRM and NAN, and retrain all these models to conduct inference without receiving target image. Our DenoiseGS shows consistent improvement across noise levels.

\begin{figure*} [t]
    \centering 
    \includegraphics[width=0.94\textwidth]{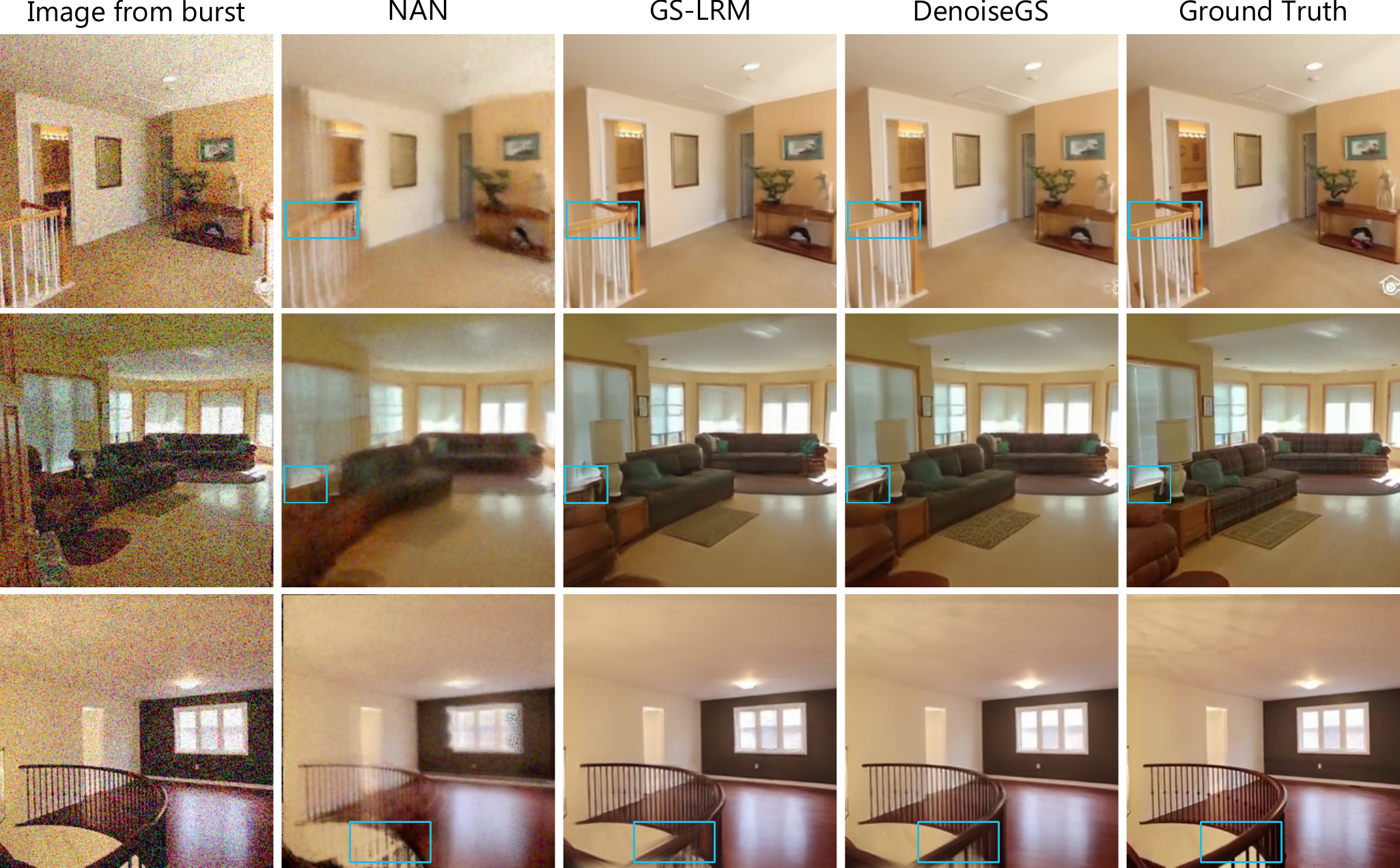}
    \vspace{-3mm}
    \caption{Qualitative comparison of novel view synthesis results under noisy conditions on the RE10K-N dataset at gain 8. Key regions are highlighted with colored rectangles. It can be observed that our method achieves superior noise suppression with less floating artifacts.} 
    \vspace{-6mm}
    \label{fig:visual_NVS}
\end{figure*}

\noindent\textbf{Qualitative Comparison on Novel View Synthesis.} Figure~\ref{fig:visual_NVS} illustrates the qualitative performance of our method on the NVS task. We observe that NAN and GS-LRM often struggle to synthesize high-quality novel views under noisy conditions. They either leave residual noise unremoved, over-smooth the images, or easily produce floaters. In contrast, our method generate novel views free of noise with fewer floaters and noticeably sharper details.

\begin{table*}[t]
    \begin{subtable}[t]{0.5\linewidth}
        \renewcommand{\arraystretch}{1.1}
        \small
        \begin{tabular}{
        c c c c |
        S[table-format=2.2]
        S[table-format=1.3]
        S[table-format=1.3]
        S[table-format=1.3]
        }
        \toprule[0.15em]
        \rowcolor{gray15}
        ~~ {GSC} ~~& ~{Warmup}~ &~ {LWF} ~&~~{RPPC}~~&~~
        PSNR$\uparrow$ ~&~~~~ SSIM$\uparrow$~~~ &~~ LPIPS$\downarrow$ ~& ~~~~{Depth$\downarrow$}~~~ \\
        \midrule[0.1em]
        - & - & - & - & 30.87 & 0.904 & 0.139 & 0.381 \\
        + & + & - & - & 31.37 & 0.906 & 0.137 & 0.210 \\
        - & - & + & - & 31.15 & 0.907 & 0.137 & 0.338 \\
        - & - & - & + & 31.09 & 0.906 & 0.140 & 0.391 \\
        + & + & + & - & 31.49 & 0.908 & 0.134 & 0.269 \\
        + & - & + & + & 31.36 & 0.902 & 0.136 & 0.261 \\
        + & + & + & + & 31.70 & 0.910 & 0.132 & 0.241 \\
        \bottomrule[0.15em]
        \end{tabular}
        \caption{Break-down ablation study}
        \label{tab:sub_a}
    \end{subtable}
    \hfill
    \begin{subtable}[t]{0.40\linewidth}
        \vspace{-20.05mm}
        \renewcommand{\arraystretch}{0.9}
        \small
        \begin{tabular}{
        l
        S[table-format=2.2]
        S[table-format=1.3]
        S[table-format=1.3]
        }
        \toprule[0.15em]
        \rowcolor{gray15}
        ~~~Method ~~&~~~~~~~ {PSNR$\uparrow$} ~~~~~~&~ {SSIM$\uparrow$} &~~~~~~~ {LPIPS$\downarrow$}~~~ \\
        \midrule[0.1em]
        ~~~Gaussian Loss & 31.39 & 0.907 & 0.138 \\
        ~~~GSC Loss      & 31.37 & 0.906 & 0.137 \\
        \bottomrule[0.15em]
        \end{tabular}
        \caption{Study on GSC Loss}
        \label{tab:sub_b}

        \vspace{0.50mm}

        \small
        \begin{tabular}{
        c
        S[table-format=2.2]
        S[table-format=1.3]
        S[table-format=1.3]
        S[table-format=2.2]
        }
        \toprule[0.15em]
        \rowcolor{gray15}
        ~~~Burst Size~~~ &~~ {PSNR$\uparrow$} ~~&~~~ {SSIM$\uparrow$} ~~&~~ {LPIPS$\downarrow$} ~~&~~~ {FPS$\uparrow$} ~~\\
        \midrule[0.1em]
        2 & 31.70 & 0.910 & 0.132 & 22.61 \\
        3 & 31.78 & 0.913 & 0.125 & 17.67 \\
        4 & 31.94 & 0.916 & 0.119 & 11.61 \\
        \bottomrule[0.15em]
        \end{tabular}
        \caption{Study on burst size}
        \label{tab:sub_c}
    \end{subtable}
    \vspace{-4mm}
    \caption{Ablations on the RE10K-N dataset. All the ablation results are evaluated at gain 8. In the break-down study (a), the "Warmup" refers to the warm-up phase for GSC loss, and the depth error is the absolute relative error. }
    \vspace{-6mm}
    \label{tab:all_ablation}
\end{table*}

\begin{figure} [t]
    \centering 
    \includegraphics[width=0.98\columnwidth]{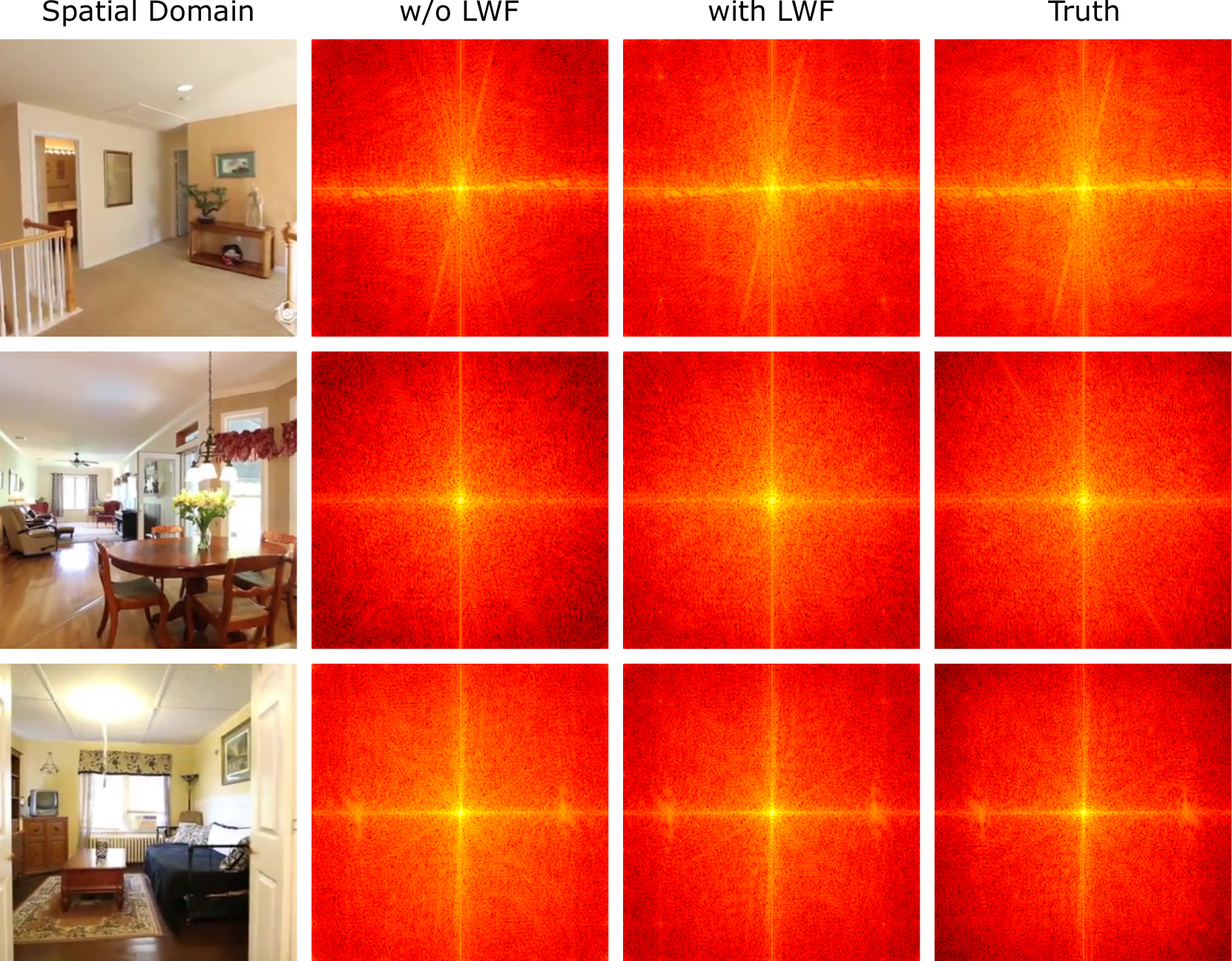}
    \vspace{-3mm}
    \caption{2D frequency spectrum visualization with and without the LWF loss. The spectra in the first two rows correspond to the G channel, while the third row corresponds to the R channel.} 
    \vspace{-7mm}
    \label{fig:freq_ablation}
\end{figure}

\vspace{-1mm}
\subsection{Ablation Study}
\label{sec:ablation}
\vspace{-1mm}

\textbf{Break-down Ablation.} In order to investigate the impact of specific design elements, we adopt the retrained GS-LRM as a baseline and conduct ablation experiments. The results are reported in Tab.~\ref{tab:sub_a}. We further compute the absolute relative depth error with respect to the depth maps generated by GS-LRM on clean inputs to assess the improvement in the spatial structure of the reconstructed Gaussian point clouds. As shown in the results, each component consistently improves performance, with the GSC loss providing the most significant gain. The use of GSC loss also contributes to the drop in the depth error. However, using GSC loss without the warm-up phase leads to a slight drop in performance, indicating the necessity of the warm-up phase.

\noindent\textbf{GSC Loss.} We conduct experiments to compare our GSC Loss with an alternative Gaussian Loss that uses GS-LRM to provide 3D supervision. Specifically, we replace the clean Gaussian point clouds in Eq.~\eqref{eq:clean_gaussians} with those generated by our retrained GS-LRM, treating them as ground-truth guidance during training. As shown in Tab.~\ref{tab:sub_b}, both approaches achieve comparable quantitative performance. However, our GSC Loss has a clear advantage in that it does not rely on any pretrained external model. It provides an intrinsic, self-consistent supervision signal generated within the same network, alleviating potential bias or domain gap.

\noindent\textbf{Frequency Spectral Visualization.} To demonstrate the advantages of our LWF loss in preserving high-frequency details, we visualize the 2D frequency spectra of the denoised results in Fig.~\ref{fig:freq_ablation}. It can be seen from the first two rows that the spectra without frequency supervision exhibit numerous dark regions around the edges, indicating the loss of high-frequency components, which will produce over-smooth artifacts in spatial domain. After applying LWF loss, these dark regions are effectively suppressed. Interestingly, as shown in the third row, we also observe that the model without LWF loss introduces spurious high-frequency components, leading to fake details in the denoised results. Our method suppresses such high-frequency components and thus produces results with more fidelity.

\noindent\textbf{Burst Size.} The burst size plays a crucial role in burst denoising. Generally, a larger burst provides more complementary information for denoising the target image but at the cost of increasing the overall computational complexity. We evaluate our model under different burst sizes in Tab.~\ref{tab:sub_c} to analyze this trade-off. To effectively handle longer bursts, we fine-tune DenoiseGS for additional 100k steps with burst sizes of 3 and 4, respectively. As shown in the table, larger bursts consistently improve restoration quality across all metrics, particularly in \text{LPIPS}. However, this gain comes at the cost of slower inference, with \text{FPS} dropping from 22.61 to 11.61. Therefore, we adopt a burst size of 2 in practice to balance quality and efficiency.

\vspace{-2mm}
\section{Conclusion}
\vspace{-2mm}

In this paper, we present DenoiseGS, the first 3DGS-based framework for burst denoising and novel view synthesis under noisy conditions. To enhance the quality of Gaussian point clouds reconstructed from noisy inputs, we propose the GSC Loss. It regularizes degraded point clouds generated under noise with the model’s own predictions from clean inputs. To better preserve fine details, we further propose the LWF Loss, which emphasizes challenging high-frequency components and provides complementary frequency-domain supervision. Extensive experiments demonstrate that our method consistently outperforms state-of-the-art approaches in both burst denoising and novel view synthesis, while maintaining fast inference speed.

{
    \small
    \bibliographystyle{ieeenat_fullname}
    \bibliography{main}

@String(CVPR= {IEEE Conf. Comput. Vis. Pattern Recog.})

@String(ICCV= {Int. Conf. Comput. Vis.})

@String(ECCV= {Eur. Conf. Comput. Vis.})

@String(TOG= {ACM Trans. Graph.})

@String(ICLR = {Int. Conf. Learn. Represent.})

@String(AAAI = {AAAI})

@String(CVPR  = {CVPR})

@String(ICCV  = {ICCV})

@String(ECCV  = {ECCV})

@String(TOG   = {ACM TOG})

@String(ICLR  = {ICLR})

@article{mildenhall2021nerf,
  title={Nerf: Representing scenes as neural radiance fields for view synthesis},
  author={Mildenhall, Ben and Srinivasan, Pratul P and Tancik, Matthew and Barron, Jonathan T and Ramamoorthi, Ravi and Ng, Ren},
  journal={Communications of the ACM},
  year={2021},
  publisher={ACM New York, NY, USA}
}

@inproceedings{mildenhall2018burst,
  title={Burst denoising with kernel prediction networks},
  author={Mildenhall, Ben and Barron, Jonathan T and Chen, Jiawen and Sharlet, Dillon and Ng, Ren and Carroll, Robert},
  booktitle={CVPR},
  year={2018}
}

@inproceedings{godard2018deep,
  title={Deep burst denoising},
  author={Godard, Cl{\'e}ment and Matzen, Kevin and Uyttendaele, Matt},
  booktitle={ECCV},
  year={2018}
}

@inproceedings{xia2020basis,
  title={Basis prediction networks for effective burst denoising with large kernels},
  author={Xia, Zhihao and Perazzi, Federico and Gharbi, Micha{\"e}l and Sunkavalli, Kalyan and Chakrabarti, Ayan},
  booktitle={CVPR},
  year={2020}
}

@inproceedings{bhat2021deep,
  title={Deep reparametrization of multi-frame super-resolution and denoising},
  author={Bhat, Goutam and Danelljan, Martin and Yu, Fisher and Van Gool, Luc and Timofte, Radu},
  booktitle={ICCV},
  year={2021}
}

@inproceedings{dudhane2022burst,
  title={Burst image restoration and enhancement},
  author={Dudhane, Akshay and Zamir, Syed Waqas and Khan, Salman and Khan, Fahad Shahbaz and Yang, Ming-Hsuan},
  booktitle={CVPR},
  year={2022}
}

@inproceedings{pearl2022nan,
  title={Nan: Noise-aware nerfs for burst-denoising},
  author={Pearl, Naama and Treibitz, Tali and Korman, Simon},
  booktitle={CVPR},
  year={2022}
}

@inproceedings{tanay2023efficient,
  title={Efficient view synthesis and 3d-based multi-frame denoising with multiplane feature representations},
  author={Tanay, Thomas and Leonardis, Ale{\v{s}} and Maggioni, Matteo},
  booktitle={CVPR},
  year={2023}
}

@inproceedings{wang2021ibrnet,
  title={Ibrnet: Learning multi-view image-based rendering},
  author={Wang, Qianqian and Wang, Zhicheng and Genova, Kyle and Srinivasan, Pratul P and Zhou, Howard and Barron, Jonathan T and Martin-Brualla, Ricardo and Snavely, Noah and Funkhouser, Thomas},
  booktitle={CVPR},
  year={2021}
}

@inproceedings{barron2021mip,
  title={Mip-nerf: A multiscale representation for anti-aliasing neural radiance fields},
  author={Barron, Jonathan T and Mildenhall, Ben and Tancik, Matthew and Hedman, Peter and Martin-Brualla, Ricardo and Srinivasan, Pratul P},
  booktitle={ICCV},
  year={2021}
}

@inproceedings{barron2022mip,
  title={Mip-nerf 360: Unbounded anti-aliased neural radiance fields},
  author={Barron, Jonathan T and Mildenhall, Ben and Verbin, Dor and Srinivasan, Pratul P and Hedman, Peter},
  booktitle={CVPR},
  year={2022}
}

@inproceedings{verbin2022ref,
  title={Ref-nerf: Structured view-dependent appearance for neural radiance fields},
  author={Verbin, Dor and Hedman, Peter and Mildenhall, Ben and Zickler, Todd and Barron, Jonathan T and Srinivasan, Pratul P},
  booktitle={CVPR},
  year={2022},
  organization={IEEE}
}

@inproceedings{hu2023tri,
  title={Tri-miprf: Tri-mip representation for efficient anti-aliasing neural radiance fields},
  author={Hu, Wenbo and Wang, Yuling and Ma, Lin and Yang, Bangbang and Gao, Lin and Liu, Xiao and Ma, Yuewen},
  booktitle={ICCV},
  year={2023}
}

@article{muller2022instant,
  title={Instant neural graphics primitives with a multiresolution hash encoding},
  author={M{\"u}ller, Thomas and Evans, Alex and Schied, Christoph and Keller, Alexander},
  journal={ACM ToG},
  year={2022},
  publisher={ACM New York, NY, USA}
}

@article{reiser2023merf,
  title={Merf: Memory-efficient radiance fields for real-time view synthesis in unbounded scenes},
  author={Reiser, Christian and Szeliski, Rick and Verbin, Dor and Srinivasan, Pratul and Mildenhall, Ben and Geiger, Andreas and Barron, Jon and Hedman, Peter},
  journal={ACM ToG},
  year={2023},
  publisher={ACM New York, NY, USA}
}

@inproceedings{chen2023mobilenerf,
  title={Mobilenerf: Exploiting the polygon rasterization pipeline for efficient neural field rendering on mobile architectures},
  author={Chen, Zhiqin and Funkhouser, Thomas and Hedman, Peter and Tagliasacchi, Andrea},
  booktitle={CVPR},
  year={2023}
}

@inproceedings{cui2024aleth,
  title={Aleth-nerf: Illumination adaptive nerf with concealing field assumption},
  author={Cui, Ziteng and Gu, Lin and Sun, Xiao and Ma, Xianzheng and Qiao, Yu and Harada, Tatsuya},
  booktitle={AAAI},
  year={2024}
}

@inproceedings{cai2024structure,
  title={Structure-aware sparse-view x-ray 3d reconstruction},
  author={Cai, Yuanhao and Wang, Jiahao and Yuille, Alan and Zhou, Zongwei and Wang, Angtian},
  booktitle={CVPR},
  year={2024}
}

@article{kerbl20233d,
  title={3D Gaussian splatting for real-time radiance field rendering.},
  author={Kerbl, Bernhard and Kopanas, Georgios and Leimk{\"u}hler, Thomas and Drettakis, George},
  journal={ACM ToG},
  year={2023}
}

@article{yang2023real,
  title={Real-time photorealistic dynamic scene representation and rendering with 4d gaussian splatting},
  author={Yang, Zeyu and Yang, Hongye and Pan, Zijie and Zhang, Li},
  journal={arXiv preprint arXiv:2310.10642},
  year={2023}
}

@inproceedings{wu20244d,
  title={4d gaussian splatting for real-time dynamic scene rendering},
  author={Wu, Guanjun and Yi, Taoran and Fang, Jiemin and Xie, Lingxi and Zhang, Xiaopeng and Wei, Wei and Liu, Wenyu and Tian, Qi and Wang, Xinggang},
  booktitle={CVPR},
  year={2024}
}

@inproceedings{luiten2024dynamic,
  title={Dynamic 3d gaussians: Tracking by persistent dynamic view synthesis},
  author={Luiten, Jonathon and Kopanas, Georgios and Leibe, Bastian and Ramanan, Deva},
  booktitle={3DV},
  year={2024},
  organization={IEEE}
}

@inproceedings{keetha2024splatam,
  title={Splatam: Splat track \& map 3d gaussians for dense rgb-d slam},
  author={Keetha, Nikhil and Karhade, Jay and Jatavallabhula, Krishna Murthy and Yang, Gengshan and Scherer, Sebastian and Ramanan, Deva and Luiten, Jonathon},
  booktitle={CVPR},
  year={2024}
}

@article{yugay2023gaussian,
  title={Gaussian-slam: Photo-realistic dense slam with gaussian splatting},
  author={Yugay, Vladimir and Li, Yue and Gevers, Theo and Oswald, Martin R},
  journal={arXiv preprint arXiv:2312.10070},
  year={2023}
}

@inproceedings{matsuki2024gaussian,
  title={Gaussian splatting slam},
  author={Matsuki, Hidenobu and Murai, Riku and Kelly, Paul HJ and Davison, Andrew J},
  booktitle={CVPR},
  year={2024}
}

@inproceedings{yan2024gs,
  title={Gs-slam: Dense visual slam with 3d gaussian splatting},
  author={Yan, Chi and Qu, Delin and Xu, Dan and Zhao, Bin and Wang, Zhigang and Wang, Dong and Li, Xuelong},
  booktitle={CVPR},
  year={2024}
}

@inproceedings{liang2024gs,
  title={Gs-ir: 3d gaussian splatting for inverse rendering},
  author={Liang, Zhihao and Zhang, Qi and Feng, Ying and Shan, Ying and Jia, Kui},
  booktitle={CVPR},
  year={2024}
}

@inproceedings{xie2024physgaussian,
  title={Physgaussian: Physics-integrated 3d gaussians for generative dynamics},
  author={Xie, Tianyi and Zong, Zeshun and Qiu, Yuxing and Li, Xuan and Feng, Yutao and Yang, Yin and Jiang, Chenfanfu},
  booktitle={CVPR},
  year={2024}
}

@inproceedings{jiang2024gaussianshader,
  title={Gaussianshader: 3d gaussian splatting with shading functions for reflective surfaces},
  author={Jiang, Yingwenqi and Tu, Jiadong and Liu, Yuan and Gao, Xifeng and Long, Xiaoxiao and Wang, Wenping and Ma, Yuexin},
  booktitle={CVPR},
  year={2024}
}

@inproceedings{liu2024humangaussian,
  title={Humangaussian: Text-driven 3d human generation with gaussian splatting},
  author={Liu, Xian and Zhan, Xiaohang and Tang, Jiaxiang and Shan, Ying and Zeng, Gang and Lin, Dahua and Liu, Xihui and Liu, Ziwei},
  booktitle={CVPR},
  year={2024}
}

@inproceedings{kocabas2024hugs,
  title={Hugs: Human gaussian splats},
  author={Kocabas, Muhammed and Chang, Jen-Hao Rick and Gabriel, James and Tuzel, Oncel and Ranjan, Anurag},
  booktitle={CVPR},
  year={2024}
}

@inproceedings{hu2024gauhuman,
  title={Gauhuman: Articulated gaussian splatting from monocular human videos},
  author={Hu, Shoukang and Hu, Tao and Liu, Ziwei},
  booktitle={CVPR},
  year={2024}
}

@article{tang2023dreamgaussian,
  title={Dreamgaussian: Generative gaussian splatting for efficient 3d content creation},
  author={Tang, Jiaxiang and Ren, Jiawei and Zhou, Hang and Liu, Ziwei and Zeng, Gang},
  journal={arXiv preprint arXiv:2309.16653},
  year={2023}
}

@inproceedings{yi2024gaussiandreamer,
  title={Gaussiandreamer: Fast generation from text to 3d gaussians by bridging 2d and 3d diffusion models},
  author={Yi, Taoran and Fang, Jiemin and Wang, Junjie and Wu, Guanjun and Xie, Lingxi and Zhang, Xiaopeng and Liu, Wenyu and Tian, Qi and Wang, Xinggang},
  booktitle={CVPR},
  year={2024}
}

@inproceedings{liang2024luciddreamer,
  title={Luciddreamer: Towards high-fidelity text-to-3d generation via interval score matching},
  author={Liang, Yixun and Yang, Xin and Lin, Jiantao and Li, Haodong and Xu, Xiaogang and Chen, Yingcong},
  booktitle={CVPR},
  year={2024}
}

@inproceedings{cai2024radiative,
  title={Radiative gaussian splatting for efficient x-ray novel view synthesis},
  author={Cai, Yuanhao and Liang, Yixun and Wang, Jiahao and Wang, Angtian and Zhang, Yulun and Yang, Xiaokang and Zhou, Zongwei and Yuille, Alan},
  booktitle={ECCV},
  year={2024},
  organization={Springer}
}

@article{zha2024r,
  title={R$^2$-Gaussian: Rectifying Radiative Gaussian Splatting for Tomographic Reconstruction},
  author={Zha, Ruyi and Lin, Tao Jun and Cai, Yuanhao and Cao, Jiwen and Zhang, Yanhao and Li, Hongdong},
  journal={arXiv preprint arXiv:2405.20693},
  year={2024}
}

@inproceedings{charatan2024pixelsplat,
  title={pixelsplat: 3d gaussian splats from image pairs for scalable generalizable 3d reconstruction},
  author={Charatan, David and Li, Sizhe Lester and Tagliasacchi, Andrea and Sitzmann, Vincent},
  booktitle={CVPR},
  year={2024}
}

@inproceedings{chen2024mvsplat,
  title={Mvsplat: Efficient 3d gaussian splatting from sparse multi-view images},
  author={Chen, Yuedong and Xu, Haofei and Zheng, Chuanxia and Zhuang, Bohan and Pollefeys, Marc and Geiger, Andreas and Cham, Tat-Jen and Cai, Jianfei},
  booktitle={ECCV},
  year={2024},
  organization={Springer}
}

@inproceedings{xu2025depthsplat,
  title={Depthsplat: Connecting gaussian splatting and depth},
  author={Xu, Haofei and Peng, Songyou and Wang, Fangjinhua and Blum, Hermann and Barath, Daniel and Geiger, Andreas and Pollefeys, Marc},
  booktitle={CVPR},
  year={2025}
}

@inproceedings{zhang2024gs,
  title={Gs-lrm: Large reconstruction model for 3d gaussian splatting},
  author={Zhang, Kai and Bi, Sai and Tan, Hao and Xiangli, Yuanbo and Zhao, Nanxuan and Sunkavalli, Kalyan and Xu, Zexiang},
  booktitle={ECCV},
  year={2024},
  organization={Springer}
}

@inproceedings{zhou2023nerflix,
  title={Nerflix: High-quality neural view synthesis by learning a degradation-driven inter-viewpoint mixer},
  author={Zhou, Kun and Li, Wenbo and Wang, Yi and Hu, Tao and Jiang, Nianjuan and Han, Xiaoguang and Lu, Jiangbo},
  booktitle={CVPR},
  year={2023}
}

@article{zhou2023nerflix++,
  title={From nerflix to nerflix++: A general nerf-agnostic restorer paradigm},
  author={Zhou, Kun and Li, Wenbo and Jiang, Nianjuan and Han, Xiaoguang and Lu, Jiangbo},
  journal={TPAMI},
  year={2023},
  publisher={IEEE}
}

@inproceedings{linhqgs,
  title={HQGS: High-Quality Novel View Synthesis with Gaussian Splatting in Degraded Scenes},
  author={Lin, Xin and Luo, Shi and Shan, Xiaojun and Zhou, Xiaoyu and Ren, Chao and Qi, Lu and Yang, Ming-Hsuan and Vasconcelos, Nuno},
  booktitle={ICLR}
}

@article{zhou2018stereo,
  title={Stereo magnification: Learning view synthesis using multiplane images},
  author={Zhou, Tinghui and Tucker, Richard and Flynn, John and Fyffe, Graham and Snavely, Noah},
  journal={arXiv preprint arXiv:1805.09817},
  year={2018}
}

@inproceedings{cai2025baking,
  title={Baking gaussian splatting into diffusion denoiser for fast and scalable single-stage image-to-3d generation and reconstruction},
  author={Cai, Yuanhao and Zhang, He and Zhang, Kai and Liang, Yixun and Ren, Mengwei and Luan, Fujun and Liu, Qing and Kim, Soo Ye and Zhang, Jianming and Zhang, Zhifei and others},
  booktitle={ICCV},
  year={2025}
}

@article{paszke2019pytorch,
  title={Pytorch: An imperative style, high-performance deep learning library},
  author={Paszke, Adam and Gross, Sam and Massa, Francisco and Lerer, Adam and Bradbury, James and Chanan, Gregory and Killeen, Trevor and Lin, Zeming and Gimelshein, Natalia and Antiga, Luca and others},
  journal={NeurIPS},
  year={2019}
}

@article{kingma2014adam,
  title={Adam: A method for stochastic optimization},
  author={Kingma, Diederik P},
  journal={arXiv preprint arXiv:1412.6980},
  year={2014}
}

@article{dao2023flashattention,
  title={Flashattention-2: Faster attention with better parallelism and work partitioning},
  author={Dao, Tri},
  journal={arXiv preprint arXiv:2307.08691},
  year={2023}
}

@misc{lefaudeux2022xformers,
  title={xformers: A modular and hackable transformer modelling library},
  author={Lefaudeux, Benjamin and Massa, Francisco and Liskovich, Diana and Xiong, Wenhan and Caggiano, Vittorio and Naren, Sean and Xu, Min and Hu, Jieru and Tintore, Marta and Zhang, Susan and others},
  year={2022}
}

@article{feng2024srgs,
  title={Srgs: Super-resolution 3d gaussian splatting},
  author={Feng, Xiang and He, Yongbo and Wang, Yubo and Yang, Yan and Li, Wen and Chen, Yifei and Kuang, Zhenzhong and Fan, Jianping and Jun, Yu and others},
  journal={arXiv preprint arXiv:2404.10318},
  year={2024}
}

@article{plucker1865xvii,
  title={Xvii. on a new geometry of space},
  author={Plucker, Julius},
  journal={Philosophical Transactions of the Royal Society of London},
  number={155},
  pages={725--791},
  year={1865},
  publisher={The Royal Society London}
}
}


\end{document}